%% file: main.tex
 \documentclass[final,5p,times,twocolumn,authoryear]{elsarticle}

\usepackage{amssymb}
\usepackage{lipsum}
\usepackage{etoolbox}
\makeatletter
\patchcmd{\ps@pprintTitle}
  {Preprint submitted}
  {To be submitted}
  {}{}
\makeatother

\usepackage{caption}
\usepackage{url}
\usepackage{hyperref}
\usepackage{multirow}
\usepackage{amsmath}
\usepackage{color}
\usepackage{wrapfig}
\usepackage{graphicx}
\setcitestyle{numbers}

\journal{Robotics and Autonomous Systems}

\begin{document}
\begin{frontmatter}



\title{Visual place recognition for aerial imagery: A survey}


\author[first,second]{Ivan Moskalenko\corref{contrib}}
\ead{I.Moskalenko@skoltech.ru}
\author[first]{Anastasiia Kornilova\corref{contrib}}
\ead{Anastasiia.Kornilova@skoltech.ru}
\author[first]{Gonzalo Ferrer}
\ead{G.Ferrer@skoltech.ru}
\cortext[contrib]{Authors contributed equally \\ \url{https://doi.org/10.1016/j.robot.2024.104837}}
\affiliation[first]{organization={Center for AI Technology (CAIT), Skolkovo Institute of Science and Technology (Skoltech)},
            city={Moscow},
            country={The Russian Federation}}
\affiliation[second]{organization={Software Engineering Department, Saint Petersburg State University},
            city={Saint Petersburg},
            country={The Russian Federation}}
            
\begin{abstract}
Aerial imagery and its direct application to visual localization is an essential problem for many Robotics and Computer Vision tasks. While Global Navigation Satellite Systems (GNSS) are the standard default solution for solving the aerial localization problem, it is subject to a number of limitations, such as, signal instability or solution unreliability that make this option not so desirable. Consequently, visual geolocalization is emerging as a viable alternative. However, adapting \emph{Visual Place Recognition} (VPR) task to aerial imagery presents significant challenges, including weather variations and repetitive patterns. Current VPR reviews largely neglect the specific context of aerial data. This paper introduces a methodology tailored for evaluating VPR techniques specifically in the domain of aerial imagery, providing a comprehensive assessment of various methods and their performance. However, we not only compare various VPR methods, but also demonstrate the importance of selecting appropriate zoom and overlap levels when constructing map tiles to achieve maximum efficiency of VPR algorithms in the case of aerial imagery. The code is available on our GitHub repository~--- \url{https://github.com/prime-slam/aero-vloc}.
\end{abstract}



\begin{keyword}
visual place recognition \sep geolocalization \sep aerial imagery \sep benchmark


\begin{center}
    \centering
    \captionsetup{type=figure}
    \includegraphics[width=0.98\textwidth]{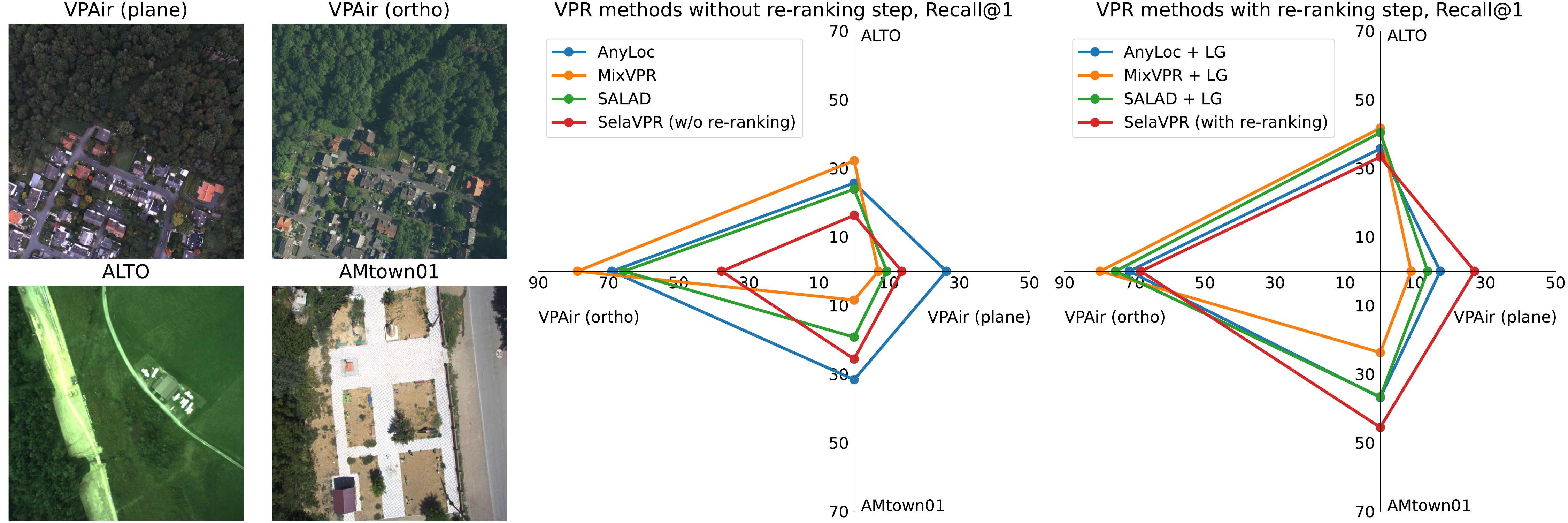}
    \captionof{figure}{Left: sample frames from the used aerial datasets for Visual Place Recognition (VPR). Center: Recall@1 statistics for VPR methods without re-ranking step. Right: Recall@1 statistics for VPR methods with re-ranking step, top-100 candidates are used everywhere.}
    \label{fig:teaser}
\end{center}
\end{keyword}
\end{frontmatter}



\input{src/000_intro}
\input{src/010_related}
\input{src/015_methodology}
\input{src/030_experiments}

\section{Conclusion}
\label{sec:conclusion}
This paper presents a comprehensive evaluation of different VPR and re-ranking methodologies tailored to aerial imagery. By introducing a novel approach for database construction, we provide a methodology for evaluating VPR methods in the context of aerial data. Furthermore, our open-source benchmark facilitates the combination and comparison of various VPR and re-ranking techniques. This paper serves as a starting point for researchers and practitioners exploring the application of aerial VPR algorithms, emphasizing the significance of hyperparameter optimization (specifically zoom and overlap) in the generation of map tiles. Additionally, it engages in discourse concerning various facets of state-of-the-art localization algorithms within the aerial domain. Moreover, the results of our study have potential applications in the deployment of visual geolocalization systems on real-world airborne platforms. While our methodology alone may not provide comprehensive robustness, it can be effectively augmented with additional sensors, such as inertial measurement units (IMUs). This integration enhances its utility for Visual Inertial Odometry (VIO) and Simultaneous Localization and Mapping (SLAM) systems, particularly for periodic location refinement and loop closure tasks. Additionally, our methodology could serve as a dependable emergency localization fallback in the event of an unexpected GNSS signal loss. We have made all contributions, including code, public.

\bibliographystyle{elsarticle-harv} 
\bibliography{example}
\vspace{10pt}
{\setlength\intextsep{0pt}
\begin{wrapfigure}{l}{25mm} 
    \includegraphics[width=1in, height=1in]{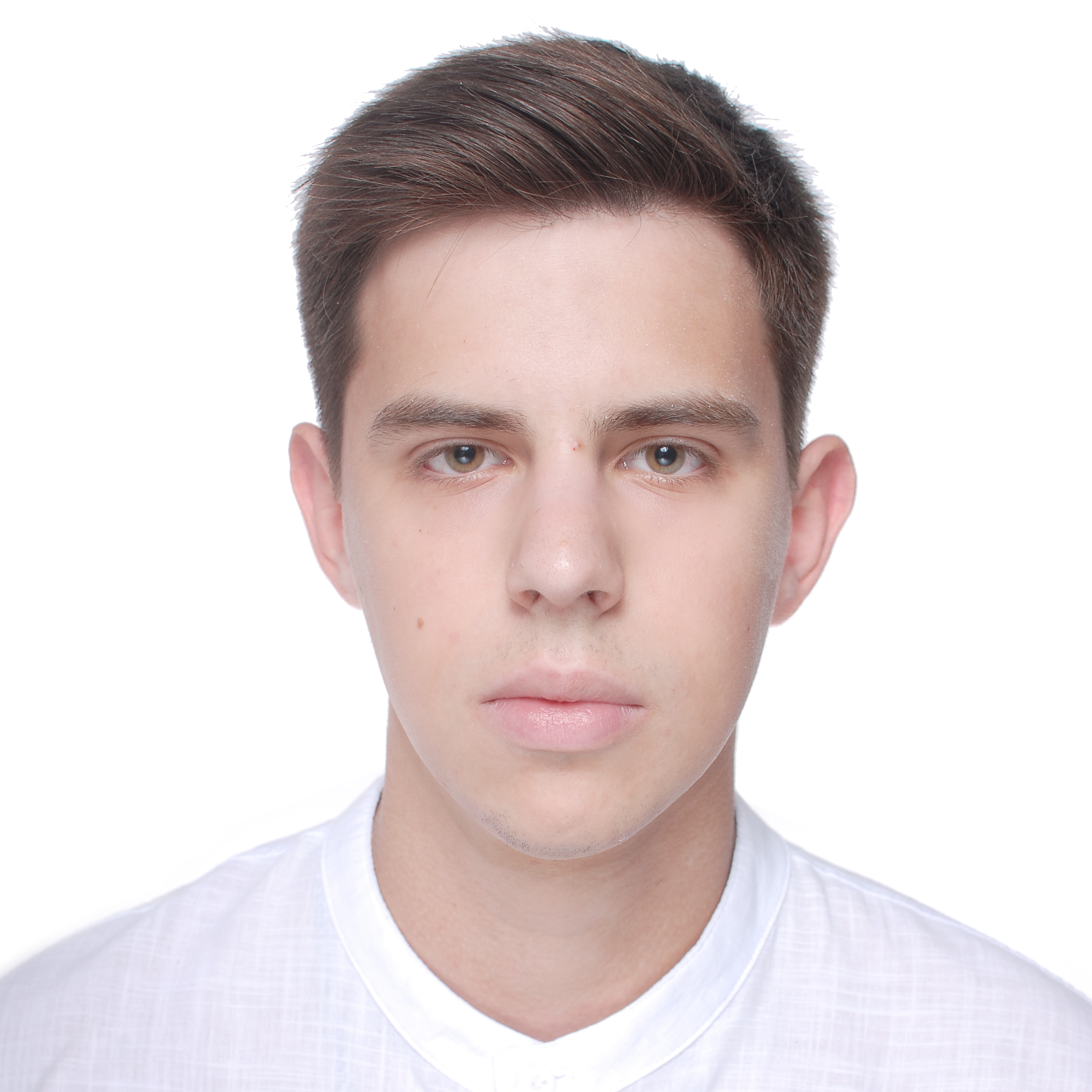}
\end{wrapfigure}\par
\noindent\textbf{Ivan Moskalenko} is pursuing a bachelor's degree at \emph{Saint Petersburg State University} in Saint Petersburg, Russia. Since 2022, he has been a Research Intern at the Mobile Robotics Lab, Skoltech, concentrating on reproducible research in SLAM and Visual Place Recognition across various data modalities.\par}
\vspace{10pt}
{\setlength\intextsep{0pt}
\begin{wrapfigure}{l}{25mm} 
    \includegraphics[width=1in, height=1in]{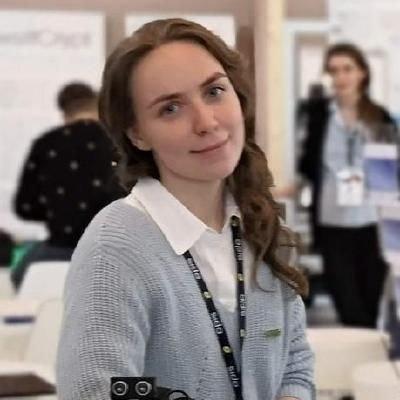}
\end{wrapfigure}\par
\noindent\textbf{Anastasiia Kornilova} obtained her B.S. (2018) degree in Software Engineering from \emph{Saint Petersburg State University}, St. Petersburg, Russia and M.S. (2021) degree in Data Science from \emph{Skolkovo Institute of Science and Technology (Skoltech)}, Moscow, Russia. 

Since 2020, Anastasiia works as a Research Engineer in Mobile Robotics Laboratory at Skoltech Center for Artificial Intelligence Technology (CAIT). Her research targets development of robust and accurate localization and SLAM methods on different visual sensor modalities. As a lab member Anastasiia leads projects with industrial partners to adapt and embed those technologies into commercial products.\par}
\vspace{10pt}
{\setlength\intextsep{0pt}
\begin{wrapfigure}{l}{25mm} 
    \includegraphics[width=1in,height=1in]{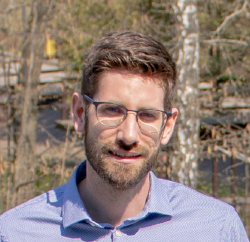}
\end{wrapfigure}\par
\noindent\textbf{Gonzalo Ferrer} obtained  his Ph.D. in Robotics from the {\em Universitat Polit\`ecnica de Catalunya} (UPC), Barcelona, Spain in 2015 and worked during two years as a Research Fellow (postdoc) at the APRIL lab. in the department of Computer Science and Engineering at the University of Michigan. In 2018, Gonzalo  joined the Skolkovo Institute of Science and Technology as an Assistant Professor. He is heading the Mobile Robotics lab., focusing his research on planning, perception and how to combine both into new solutions in robotics.\par}
\end{document}

%% file: src/000_intro.tex
\section{Introduction}
Aerial imagery has become an indispensable tool across various sectors, including geology~\cite{umar20163d}, agriculture~\cite{rejeb2022drones}, environmental monitoring~\cite{mastelic2020aerial}, disaster management~\cite{khan2022emerging}, and civil engineering~\cite{lu2024accurate}. The effectiveness of aerial imagery, however, is contingent on accurate localization, which involves determining the geographical coordinates of the images. One method for localizing aerial imagery is visual geolocalization, which identifies the location of a captured image using only visual information.

Global Navigation Satellite Systems (GNSS), such as GPS and GLONASS, are commonly employed to localize aerial imagery. However, this method requires stable satellite signals and absence of interference, conditions that are not always achievable, especially in the presence of signal reflections from water bodies or complex terrain. Additionally, the uneven signal coverage of global satellite systems introduces further challenges, thus enhancing the relevance of visual geolocalization.

Visual geolocalization can be implemented through various methods, typically relying on a pre-built database of images with known locations~\cite{lowry2015visual}. This approach generally involves two stages: global localization (or \emph{Visual Place Recognition}, VPR) and local alignment. Global localization involves identifying the nearest frame from the database (\emph{Image Retrieval}), while local alignment determines the precise position using the selected frame. VPR is often achieved using global descriptors, which are compact representations of images derived from aggregated image features. These compact representations enable rapid searching even in extensive databases~\cite{douze2024faiss}, though they suffer from relatively low accuracy, which necessitates the use of re-ranking approaches to enhance quality~\cite{barbarani2023local}, albeit at the cost of increased processing time.

The application of Visual Place Recognition methods to aerial imagery is complicated by several factors, such as varying weather conditions, changing seasons, and different times of day. Aerial image localization requires sufficiently large maps, imposing additional constraints on the global localization stage. Scale variations due to differences in altitude further complicate global localization. For these systems to function effectively, the database images must closely match the scale of the query images. Additionally, aerial images often feature repetitive patterns, such as urban grids and agricultural fields, making localization based on visual information challenging. These factors present significant challenges for contemporary VPR methods.

To date, reviews and comparisons of VPR systems have primarily focused on indoor and outdoor data, neglecting satellite and aerial imagery~\cite{masone2021survey, garg2021your, zaffar2020visual, berton2022deep}. This paper aims to deliver a thorough comparative analysis of current VPR and re-ranking methodologies in the context of aerial imagery, as well as to identify the most effective map representation for this application. To achieve this, we introduce a novel methodology for constructing databases for the VPR task, which encompasses the determination of the optimal scale level and the level of overlap between adjacent map tiles. Additionally, we present an open-source benchmark that incorporates a range of popular VPR and re-ranking techniques.

The main contributions of the paper are as follows:
\begin{itemize}
    \item a methodology for evaluating VPR methods in case of aerial data, including database construction part;
    \item an open-source benchmark that allows combining various VPR and re-ranking techniques;
    \item a complete evaluation of different VPR methods, providing both quality and performance comparisons.
\end{itemize}

The paper is structured to address the following components: \hyperref[sec:rw]{(Section 2)} reviews related work on VPR methods, re-ranking techniques, and datasets for aerial imagery. \hyperref[sec:methodology]{(Section 3)} describes the methodology, including database construction and evaluation metrics. \hyperref[sec:experiments]{(Section 4)} and \hyperref[sec:conclusion]{(Section 5)} cover the experimental setup, results, and discussion.

%% file: src/010_related.tex
\section{Related Work}
\label{sec:rw}
This section provides a comprehensive review of existing \emph{Visual Place Recognition} systems, highlighting various re-ranking techniques designed to enhance their efficacy~\cite{barbarani2023local}. The summary of this review is provided in Table~\ref{tab:vpr_review}. It particularly emphasizes on the review articles pertinent to this topic. Special consideration is given to the VPR in aerial imagery, detailing commonly utilized datasets. Additionally, this section discusses the overview of evaluation technologies employed in assessing the performance of VPR methodologies.

\begin{table*}
    \centering
    \begin{tabular}{ccccc}
         Ref & Method & Year & Backbone & Test domains \\
         \hline \hline
         \cite{cummins2008fab} & FAB-MAP & 2008 & --- & Urban \\
         \hline
         \cite{galvez2012bags} & Gálvez-López et al. & 2012 & --- & Indoor, Urban \\
         \hline
         \cite{arandjelovic2016netvlad} & NetVLAD & 2016 & AlexNet, VGG-16 & Urban \\
         \hline
         \cite{shetty2019uav} & Shetty et al. & 2019 & AlexNet & Aerial \\
         \hline
         \cite{mantelli2019novel} & Mantelli et al. & 2019 & --- & Aerial \\
         \hline
         \cite{zaffar2020cohog} & CoHOG & 2020 & --- & Indoor, Urban \\
         \hline
         \cite{zhuang2021faster} & Zhuang et al. & 2021 & ResNet & Aerial \\
         \hline
         \cite{dai2021transformer} & Dai et al. & 2021 & ViT-S & Aerial \\
         \hline
         \cite{bianchi2021uav} & Bianchi et al. & 2021 & Custom autoencoder & Aerial \\
         \hline
         \cite{berton2022rethinking} & CosPlace & 2022 & VGG-16, ResNet & Urban \\
         \hline
         \cite{berton2023eigenplaces} & Eigenplaces & 2023 & VGG-16, ResNet & Urban \\
         \hline
         \cite{ali2023mixvpr} & MixVPR & 2023 & ResNet & Urban \\
         \hline
         \multirow{2}{*}{\cite{keetha2023anyloc}} & \multirow{2}{*}{AnyLoc} & \multirow{2}{*}{2023} & \multirow{2}{*}{DINOv2} & Indoor, Urban, Underwater, \\ 
         &&&& Aerial, Subterranean \\
         \hline
         \cite{izquierdo2023optimal} & SALAD & 2023 & DINOv2 & Urban \\
         \hline
         \cite{wang2024multiple} & Wang et al. & 2024 & ResNet & Aerial \\
         \hline
         \cite{lu2024towards} & SelaVPR & 2024 & DINOv2 & Urban \\
    \end{tabular}
    \caption{Overview of existing \emph{Visual Place Recognition} systems}
    \label{tab:vpr_review}
\end{table*}

\subsection{Visual Place Recognition methods}
The goal of VPR algorithms is to find the most appropriate image among a database of images and the locations that match them. This objective constitutes a subset within the broader domain of \emph{Image Retrieval}~\cite{datta2008image, wan2014deep}, with applications spanning search engines, visual content analysis, and robotics. The classic approach for \emph{Visual Place Recognition}, as employed in early works~\cite{cummins2008fab, galvez2012bags}, involves utilizing descriptors, which are compact representations of images. They are obtained by aggregating local features of images into a vector of fixed length. Then, L2 or cosine similarity is used to search in the database of descriptors.

During the nascent phases of feature aggregation methodology, the bag-of-words~\cite{galvez2012bags} technique gained significant traction. Subsequent to this, the scientific community was acquainted with the VLAD (Vector of Locally Aggregated Descriptors)~\cite{arandjelovic2013all} methodology, which entails the fusion of local descriptors obtained from keypoints within an image into a vectorized form via the calculation of disparities between each descriptor and a set of predefined visual clusters. Following this, a trainable adaptation of VLAD, denoted as NetVLAD~\cite{arandjelovic2016netvlad}, was introduced to the community. NetVLAD utilizes AlexNet~\cite{krizhevsky2012imagenet} and VGG-16~\cite{simonyan2014very} as base architectures. However, methods using hand-crafted features have also seen development. For example, Zaffar et al.~\cite{zaffar2020cohog} employ Histogram-of-Oriented-Gradients (HOG) descriptors and image entropy to extract regions of interest and perform regional-convolutional descriptor matching. More contemporary VPR techniques like CosPlace~\cite{berton2022rethinking} and its viewpoint-robust variant, EigenPlaces~\cite{berton2023eigenplaces}, are evaluated on both VGG-16 and the newer backbone ResNet~\cite{he2016deep}, with the latter demonstrating superior performance. MixVPR~\cite{ali2023mixvpr} adopts ResNet as its backbone and employs the all-MLP feature aggregation technique.

Foundation models, such as CLIP~\cite{radford2021learning}, mark a significant shift in AI by providing models with a profound grasp of text and images. Through pre-training on extensive text-image datasets, these models redefine the landscape of image processing AI. Notably, recent remarkable achievements in the VPR task have also been realized with foundation models, namely DINOv2~\cite{oquab2023dinov2}. The pioneering approach that utilizes DINOv2 is the method proposed by Keetha et al. and known as AnyLoc~\cite{keetha2023anyloc}. They demonstrate that using foundation models without VPR-specific training is able to show SOTA quality. AnyLoc also provides a comparison of their method with competing methods on two aerial-to-aerial datasets. The idea of using foundation models has been developed in the SALAD~\cite{izquierdo2023optimal} and SelaVPR~\cite{lu2024towards} approaches. The first approach proposes a new aggregation technique based on the optimal transport problem. SelaVPR, on the other hand, proposes a method to realize Seamless adaptation by adding a few tunable lightweight adapters to the frozen pre-trained model.

One of the principal drawbacks of numerous works in the field of \emph{Visual Place Recognition} is the insufficient attention given to the performance issues of the proposed methods. Concurrently, there is a trend towards utilizing large foundation models, such as DINOv2. For instance, AnyLoc employs the largest DINOv2 model in terms of parameters—ViT-G. Although this model enhances prediction quality, the time required for prediction can be excessively long, even on high-performance devices. Therefore, it is anticipated that future research in VPR will increasingly focus on addressing performance issues, particularly in the context of embedded devices.

\subsection{Visual Place Recognition surveys}
Over the past few years, a number of surveys exploring different approaches to \emph{Visual Place Recognition} have become available to the community. The purpose of some~\cite{barros2021place, masone2021survey} is only to review different approaches to VPR or to find possible applications~\cite{garg2021your}, and no data-driven comparisons are made. 

Zaffar et al.~\cite{zaffar2021vpr} offer a tool to the community to compare different VPR methods and present their experimental comparison, but it is only done on indoor and outdoor datasets and does not cover the aerial case. The same can be said for the study by Pion et al.~\cite{pion2020benchmarking}.

Wilson et al.~\cite{wilson2024image} talk about different approaches to visual geolocalization, including cross-view geolocalization, where the query image is a ground-view image and satellite images are used in the VPR database. However, they do not focus on aerial-to-aerial localization and do not conduct an experimental study on them.

Berton et al.~\cite{berton2022deep} in their paper provide a comparison of different backbones and aggregation techniques for the VPR task, but they also use only indoor and outdoor datasets for this purpose.

Li et al.~\cite{li2023evaluation} evaluate fifteen global descriptor methods for VPR by analyzing their performance across six diverse datasets, encompassing indoor environments, urban roads, suburban areas, and natural scenery. The study provides design recommendations for enhancing global descriptors and examines the trade-offs between matching performance and computational efficiency in practical VPR applications. However, this study does not address the application of VPR techniques in the context of aerial imagery.

Also worth mentioning in this section is the paper by Barbarani et al.~\cite{barbarani2023local} In it, they evaluate different re-ranking methods that are actively used in VPR approaches. However, this study also focuses on indoor and outdoor datasets and does not address aerial data.

\subsection{Re-ranking methods}
Despite the rapid variety of techniques for local feature extraction and their further aggregation, the application of various re-ranking methods remains relevant~\cite{barbarani2023local}. This holds particular significance in the context of aerial imagery, where there is often a high degree of visual homogeneity observed across urban, semi-urban, and especially pristine natural landscapes. In essence, re-ranking methods are also \emph{Visual Place Recognition} methods with the difference that local features are used directly, without their prior aggregation, to determine the similarity measure between two images. These methods show higher quality but also require more inference time. Therefore, usually re-ranking method is fed with N candidate images obtained by classical VPR methods, from which it selects K best images.

One approach for re-ranking is to use local features such as keypoints with their corresponding descriptors. Then for re-ranking, keypoints of the query image are matched with sets of keypoints of candidate images, and the criterion is the number of matched keypoints. Both RANSAC~\cite{fischler1981random} and the more advanced SuperGlue~\cite{sarlin2020superglue} or LightGlue~\cite{lindenberger2023lightglue} trainable methods can be used for keypoint matching. While this approach isn't specifically tailored for the Image Retrieval task, it is used by the community.

Another approach for the re-ranking task is to use dense local features. For example, Patch-NetVLAD~\cite{hausler2021patch} applies VLAD~\cite{arandjelovic2013all} aggregation to extract image patch descriptors and match them using RANSAC or Rapid Scoring. Recent SOTA solutions in this area are ETR~\cite{zhang2023etr}, R2Former~\cite{zhu2023r2former} and SelaVPR~\cite{lu2024towards}. These methods use ViT backbones to extract dense local features both for their matching directly and for further aggregation. Thus, these solutions combine both classical Image Retrieval approach and re-ranking.

\subsection{VPR in aerial images}
\emph{Visual Place Recognition} for aerial imagery primarily emerges in research works centered on UAV localization, with various studies exploring advanced methodologies to improve image matching capabilities between UAV-captured and satellite images.

Multiple research efforts have focused on adapting existing deep learning architectures to enhance their effectiveness in this application. For example, Shetty et al.~\cite{shetty2019uav} propose using the AlexNet architecture without its final classification layer, while Zhuang et al.~\cite{zhuang2021faster} recommend modifications to the ResNet-50 architecture.

Additionally, uses of other models have been highlighted. Dai et al.~\cite{dai2021transformer} suggest employing a small Visual Transformer (ViT-S), pre-trained on ImageNet, as a robust backbone for their matching tasks. Similarly, Bianchi et al.~\cite{bianchi2021uav} investigate the potential of an autoencoder architecture for improving UAV and satellite image matching.

In the case of Mantelli et al.~\cite{mantelli2019novel}, the abBRIEF descriptor is utilized to calculate the similarity score of images. In contrast, both Chen et al.~\cite{chen2021real} and Hao et al.~\cite{hao2023range} integrate the NetVLAD algorithm for global localization but employ SuperGlue for re-ranking. Moreover, Gurgu et al.~\cite{gurgu2022vision} exclusively use SuperGlue for selecting the best matching satellite image, which imposes limitations on the number of map tiles.

Wang et al.~\cite{wang2024multiple} identify weather and lighting variations as significant challenges for visual geo-localization systems. To mitigate these issues, they propose MuSe-Net, an adaptive learning framework incorporating a dual-path CNN to minimize environmental style discrepancies and a Residual SPADE module to enhance feature discrimination and optimize training.

Despite the existence of a variety of \emph{Visual Place Recognition} solutions tailored for aerial imagery, this paper primarily concentrates on classical VPR methodologies.

\subsection{Datasets}
Currently, the availability of publicly accessible datasets comprising aerial imagery is relatively limited. Among these, the University-1652~\cite{zheng2020university} dataset is notable, encompassing images of 1,652 buildings distributed across 72 universities globally. Additionally, the CVUSA~\cite{workman2015wide} dataset is widely utilized, containing several million photographs captured throughout the United States. Another significant dataset is SUES-200~\cite{zhu2023sues}. The datasets such as DenseUAV~\cite{dai2023vision}, ALTO~\cite{cisneros2022alto}, VPAir~\cite{schleiss2022vpair}, and MARS-LVIG~\cite{li2024mars} consist of sequential images captured from aircraft and are employed in the assessment of visual geolocalization algorithms. Furthermore, notable contributions in this field include the works of Tian et al.~\cite{tian2017cross} and Lin et al.~\cite{lin2015learning}, along with datasets VIGOR~\cite{zhu2021vigor} and DAG~\cite{vallone2022danish}.

%% file: src/015_methodology.tex
\begin{figure*}
    \centering
    \captionsetup{type=figure}
    \includegraphics[width=1\textwidth]{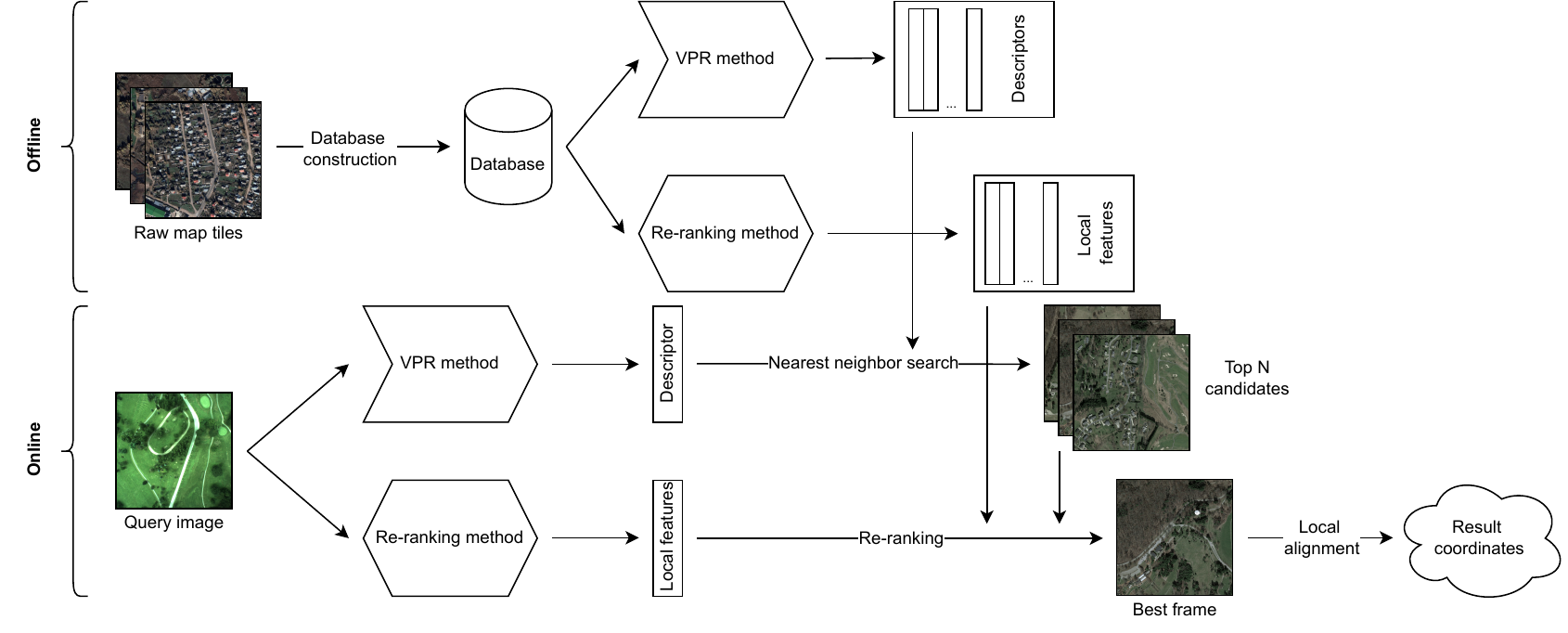}
    \captionof{figure}{Pipeline of the proposed aerial visual geo-localization system. The offline phase involves the computation of global descriptors and local features for map tiles. During the online phase, the VPR method selects the $N$ nearest images from the database, a re-ranking method identifies the optimal frame from the candidate set, and the precise location is determined through local alignment.}
    \label{fig:pipeline}
\end{figure*}

\section{Methodology}
\label{sec:methodology}
The suggested approach adheres to a typical framework for Visual Place Recognition, with its comprehensive workflow depicted in Figure~\ref{fig:pipeline}. The process is divided into two main segments: offline and online. The offline segment involves building a database for VPR techniques. The online segment includes the \emph{Image Retrieval} component as well as \emph{Local Alignment}, which determines the precise position of the system using the chosen frame. Although our approach shares similarities with other VPR benchmarks, it incorporates several unique innovations. Although the challenge of determining appropriate zoom and overlap levels between frames in database construction from aerial imagery has been previously highlighted in the academic community~\cite{dai2023vision, hao2023range}, we introduce a generalized parameterization method to address this problem. Additionally, we propose two new metrics that are better suited for aerial imagery compared to the widely used \emph{Recall@k} metric. 

In this section, we elaborate on the methodologies employed in the construction of the database~\hyperref[sec:database]{(Section A)}, as well as the procedures involved in \emph{Image Retrieval}~\hyperref[sec:retrieval]{(Section B)} and \emph{Local Alignment}~\hyperref[sec:local_alignment]{(Section C)} within the proposed pipeline. Additionally, the criteria for the selection of test datasets are detailed~\hyperref[sec:datasets]{(Section D)}. This section also introduces two novel metrics developed for the assessment of VPR methods in the context of aerial imagery~\hyperref[sec:metrics]{(Section E)}.

\subsection{Database construction}
\label{sec:database}
This paper focuses on the use of satellite imagery as a reference, however, the proposed approach for database construction can be used with any data that has continuous coverage of the area over which visual geolocalization is planned.

The main idea of the proposed methodology for database construction is to partition the whole space into rectangular tiles of equal size. Such a set of tiles can already be used as a database for VPR methods~\cite{gurgu2022vision}. Nevertheless, this approach has some disadvantages, so we propose some improvements for it.

\begin{figure}
    \centering
    \captionsetup{type=figure}
    \includegraphics[width=0.45\textwidth]{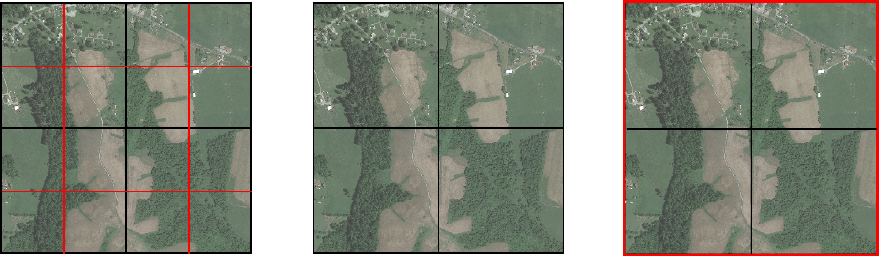}
    \captionof{figure}{Different zoom levels. Black means raw tiles, red means constructed tiles. Left: zoom 200\%. Center: zoom 100\%. Right: zoom 50\%.}
    \label{fig:zoom_levels}
\end{figure}

The first problem is that typically open data sources, such as Google Maps or Sentinel, offer a choice of only a few levels of map zoom level. Thus, even between neighboring levels, the difference can be significant to affect the performance quality of VPR methods. Therefore, we propose to not only choose the most appropriate zoom level when downloading images, but also to construct images of different scales locally using raw images. In this case, we consider that the zoom level of the downloaded raw images is equal to 100\%. An example of constructing map tiles of different zoom levels can be seen in Fig.~\ref{fig:zoom_levels}.

\begin{figure}
    \centering
    \captionsetup{type=figure}
    \includegraphics[width=0.5\textwidth]{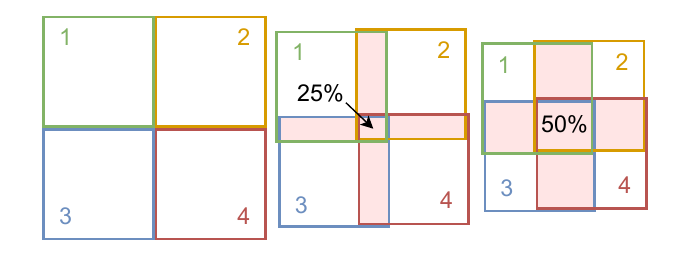}
    \captionof{figure}{Different overlap levels. The digits represent the numbering of the map tiles. Left: overlap 0\%. Center: overlap 25\%. Right: overlap 50\%.}
    \label{fig:overlap_levels}
\end{figure}

The second problem with the naive approach is the presence of stitches between tiles in the map. Thus, if a query image is captured somewhere between two adjacent tiles of the map, the probability of its successful localization is significantly reduced. To offset this problem, we propose to generate overlapping tiles. An example of database construction from images with different overlap levels between frames can be seen in Fig.~\ref{fig:overlap_levels}.

In our proposed framework, the capability to generate maps with arbitrary overlap and zoom levels is implemented using the OpenCV~\cite{opencv_library} library. The tool requires only a pre-downloaded map with zero overlap and any specified zoom level. A script for downloading maps using the Google Maps API is also available within our framework.

\subsection{Image retrieval}
\label{sec:retrieval}
Following the standard procedure for the Visual Place Recognition (VPR) task, we utilize pre-computed descriptors of map tiles and local features required for the re-ranking step. The query frame localization process is as follows: for each frame, its descriptor is computed using the VPR method. In our study, we employed the authors' implementations based on PyTorch~\cite{Ansel_PyTorch_2_Faster_2024}. The database is then searched using these descriptors, and the $N$ nearest frames are identified. Our tool leverages the Faiss~\cite{douze2024faiss} library for efficient descriptor database search. Subsequently, among the $N$ candidate frames, $K$ best matches are selected using re-ranking algorithms.

\subsection{Local alignment}
\label{sec:local_alignment}
The task of \emph{Local Alignment} is to calculate the geo-coordinates of the query image using the previously found corresponding satellite map tile. To do this, we use the keypoints of the query and satellite images and match them. Then we compute the perspective transformation, which allows us to map the query image to the satellite map tile. Since the satellite image itself is a rectangle with previously known geo-coordinates of its corners, it is possible to convert pixel coordinates to geographic coordinates. For this purpose, our approach provides Web Mercator model, which is used in Google Maps. Functionality for local alignment is implemented in the proposed framework using the corresponding methods \texttt{cv2.findHomography} and \texttt{cv2.perspectiveTransform} from the OpenCV library~\cite{opencv_library}.

\subsection{Datasets}
\label{sec:datasets}
Since one of the novelties of our approach is the database construction algorithm, our primary criterion for test datasets is that the images originate from a delimited geographical area, enabling the acquisition of corresponding satellite maps. Additionally, we require precise geocoordinates for these test images. Since our research concentrates on aerial-to-satellite localization, datasets generated from satellite imagery were not considered. Consequently, the VPAir~\cite{schleiss2022vpair}, ALTO~\cite{cisneros2022alto}, MARS-LVIG~\cite{li2024mars} and DenseUAV~\cite{dai2023vision} datasets align with our requirements. 

The VPAir~\cite{schleiss2022vpair} dataset consists of two sequences: plane and ortho. The plane sequence is captured from an aircraft, while the ortho sequence comprises orthophotos derived from aerial data. Both sequences are utilized as test sequences in our experiments. Although the ALTO~\cite{cisneros2022alto} dataset offers two sequences (Round1 and Round2), the former is a subset of the latter, prompting our utilization of Round2 exclusively. As for the MARS-LVIG~\cite{li2024mars} dataset, it encompasses various sequences; we opted for the AMtown01 sequence, one of the longest, captured over rural towns. DenseUAV~\cite{dai2023vision} is shot in similar conditions to AMtown01 in terms of altitude (80~--- 100 meters) and landscape, so we did not use it, favoring AMtown01. The characteristics of the test sequences used are given in Tab.~\ref{tab:datasets}.

\begin{table}
    \centering
    \begin{tabular}{c|c|c|c}
        \multirow{2}{*}{Dataset} & Number of & Trajectory & Resolution of\\
        ~ & frames & length & frames\\
        \hline
        VPAir (plane) & 2706 & 100 km & 800x600\\
        VPAir (ortho) & 2706 & 100 km & 800x600\\
        ALTO & 460 & 37 km & 500x500\\
        AMtown01 & 2620 & 4.8 km & 2448x2048\\
    \end{tabular}
    \caption{Characteristics of the test datasets used}
    \label{tab:datasets}
\end{table}

\subsection{Metrics}
\label{sec:metrics}
In the visual place recognition task, the standard metric is \emph{Recall@k}~\cite{berton2022deep, arandjelovic2016netvlad, berton2021viewpoint, berton2021adaptive, hausler2021patch, jin2017learned, liu2019stochastic, peng2021semantic, peng2021attentional, warburg2020mapillary}, which measures the percentage of query images for which at least one of the $K$ best images selected by the method is no further away than a given threshold. This metric assumes that the geo-coordinates of the images retrieved from the database are utilized as the resultant geo-coordinates. However, within the realm of aerial data, this approach isn't appropriate due to the substantial size of map tiles, raising ambiguity regarding the association of specific geo-coordinates with images from the database. To address this challenge, two adaptations of the original \emph{Recall@k} metric have been suggested. One assesses the complete visual geolocalization process, encompassing the local alignment stage, while the other focuses solely on the global localization phase. It's worth noting that for both suggested metrics, we assume that the query images are nadir images, and we consider their 2D projection. Consequently, we presume that the coordinates of the query image represent the latitude and longitude of its center.

Before presenting the metrics, we introduce the following general notations:
\begin{itemize}
    \item{$DB$~--- the set of map tiles in the database;}
    \item{$Q$ and $q$~--- a sequence of query images and a single query image, respectively;}
    \item{$q_{loc_{gt}}$~--- the ground-truth coordinates of the query image;}
    \item{$VPR(q, DB, N)$~--- a function representing a VPR pipeline, which may include a re-ranking step, that finds the N closest images in the database for a given query image;}
    \item{$DB_k = VPR(q, DB, k)$.}
\end{itemize}

\begin{figure}
    \centering
    \captionsetup{type=figure}
    \includegraphics[width=0.55\textwidth]{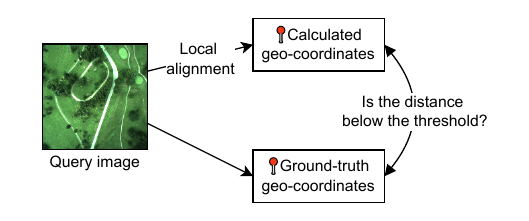}
    \captionof{figure}{A schematic representation of the {\bf Georeference Recall} metric}
    \label{fig:gr_recall}
\end{figure}

The first proposed metric is {\bf Georeference Recall}. It calculates the locations of the query images and calculates the distances to the ground-truth coordinates. If the distance is less than the specified threshold value, this image is considered to be correctly localized. The result of the metric is the ratio of correctly localized pairs to their total number. Thus, the metric allows to assess the accuracy of georeferencing, that is, the entire visual geolocalization pipeline. In essence, this metric is similar to \emph{Recall@1}, with the only difference that the resulting location is not the geo-coordinates of the selected map tile, but the results of local alignment relative to it. Its formal definition is as follows:

\begin{gather}
R_{georef.} = \frac{\sum_{q \in Q}
\begin{cases}
      1 & \text{if $ dist(q_{loc_{gt}}, LA(q, DB_{1})) < \mu$} \\
      0 & \text{else}
\end{cases}}{|Q|},
\end{gather}
where $LA$ is a local alignment function that finds the exact location of the system, $dist$ is the geographic distance function and $\mu$ is a threshold value. A schematic representation of the metric is available in Fig.~\ref{fig:gr_recall}.

\begin{figure}
    \centering
    \captionsetup{type=figure}
    \includegraphics[width=0.45\textwidth]{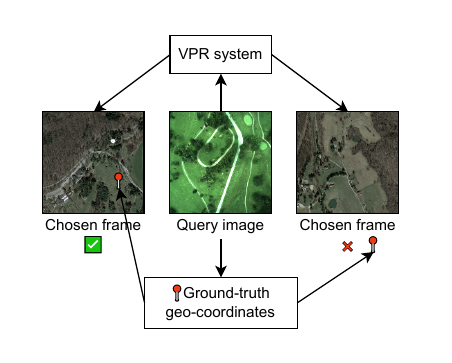}
    \captionof{figure}{A visual depiction of the {\bf VPR Recall} metric}
    \label{fig:vpr_recall}
\end{figure}

Another suggested metric is {\bf VPR Recall}, which enables the evaluation of global localization systems. Due to the database images being flat rectangular map tiles, only their corner coordinates are known by default. Therefore, this metric computes $N$ best predictions from the image database using a global localization system for each query image, and then if the center of the query image hits at least one of the selected images from the database, this result is considered as correct. The result of the metric is the ratio of correct results to the total number of test images. Thus, the metric allows to evaluate the quality of global localization systems, does not require specifying a specific threshold value and can work with different zoom levels of satellite maps. Its formal definition is as follows:

\begin{gather}
R_{VPR} = \frac{\sum_{q \in Q}
\begin{cases}
      1 & \text{if $\exists d \in DB_{N}$ s.t. $q_{loc_{gt}} \in d_{rect}$} \\
      0 & \text{else}
\end{cases}}{|Q|},
\end{gather}
where $d_{rect} = (d_{NW}, d_{SW}) \times (d_{NW}, d_{NE})$ and $d_{NW}$, $d_{SW}$, $d_{NE}$ represent the corresponding corner coordinates of the map tile. A visual depiction of the metric is presented in Fig.~\ref{fig:vpr_recall}.

%% file: src/030_experiments.tex
\section{Experiments}
\label{sec:experiments}
In this section, we perform a qualitative comparative analysis of different VPR methods, examining their performance under different zoom levels and different levels of overlap between neighboring tiles. Furthermore, we present the results of our proposed local alignment method. Additionally, this section provides a summary of the comparative analysis of the different VPR methods in terms of both time efficiency and memory usage.

\subsection{Test bench}
During our experiments, we employed four test sequences: ALTO~\cite{cisneros2022alto}, AMtown01~\cite{li2024mars}, as well as two sequences from the VPAir~\cite{schleiss2022vpair} dataset (plane and ortho). Three maps were downloaded from Google Maps to encompass the entirety of the test trajectories. From the variety of zoom levels offered by the Google Maps API, we opted for zoom level 17 for ALTO and VPAir, and zoom level 19 for AMtown01, chosen to correspond closely with the scale of the test data. In terms of our methodology, these maps were considered as maps with a zoom level equal to 100\%. Detailed specifications of the maps are provided in Table~\ref{tab:maps}.

In addressing the VPR task, we employed several prominent methods that have emerged in recent years: CosPlace~\cite{berton2022rethinking}, EigenPlaces~\cite{berton2023eigenplaces}, MixVPR~\cite{ali2023mixvpr}, and SALAD~\cite{izquierdo2023optimal}, alongside the established NetVLAD~\cite{arandjelovic2016netvlad} approach. The original weights provided by the respective authors were utilized without any fine-tuning. For the re-ranking phase, two methods utilizing keypoints were used: the conventional SuperGlue~\cite{sarlin2020superglue} approach and its recent variant, LightGlue~\cite{lindenberger2023lightglue}. Additionally, we incorporated the novel SelaVPR~\cite{lu2024towards} method, which has demonstrated superior performance on several popular VPR datasets. SelaVPR integrates both candidate selection for the re-ranking phase and the re-ranking process itself.

\begin{table}
    \centering
    \begin{tabular}{c|c|c|c}
        \multirow{2}{*}{Dataset} & Number of & \multirow{2}{*}{Area} & Resolution of \\
        & raw tiles & ~ & tiles \\
        \hline
        VPAir & 4788 & 1079 km$^2$ & 1280x1230\\
        ALTO & 2706 & 718 km$^2$ & 1280x1230\\
        AMtown01 & 3190 & 66 km$^2$ & 1280x1230\\
    \end{tabular}
    \caption{Characteristics of the maps used}
    \label{tab:maps}
\end{table}

\subsection{Quality evaluation}
\subsubsection{Zoom level}
To compare VPR methods across various zoom levels of the map, multiple databases were constructed with an overlap level of 25\%. All considered VPR methods were utilized. The zoom levels employed were 50\%, 100\%, 150\%, and 200\%. Additionally, for the ALTO~\cite{cisneros2022alto} dataset, a zoom level of 250\% was included, as multiple methods demonstrated the best performance at the 200\% zoom level. The {\bf{VPR Recall}} metric was used. The results of the comparison are summarized in Table~\ref{tab:zoom}.

\begin{table*}
    \centering
    \resizebox{\textwidth}{!}{
    \begin{tabular}{l|l|c|c|c|c|c|c|c|c|c|c|c|c|c|c|c|c}
        \multirow{2}{*}{Dataset} & Localization & \multicolumn{3}{c|}{Zoom 50\%} & \multicolumn{3}{c|}{Zoom 100\%} & \multicolumn{3}{c|}{Zoom 150\%} & \multicolumn{3}{c|}{Zoom 200\%} & \multicolumn{3}{c|}{Zoom 250\%} & Best \\\cline{3-17}
        & method & R@1 & R@10 & R@50 & R@1 & R@10 & R@50 & R@1 & R@10 & R@50 & R@1 & R@10 & R@50 & R@1 & R@10 & R@50 & zoom \\
        \hline \hline
        \multirow{7}{*}{VPAir (plane)} & \textcolor{green}{AnyLoc} & 5.1 & 22.8 & 54.4 & 14.2 & 36.0 & 57.4 & \textbf{21.7} & \textbf{43.6} & \textbf{61.2} & 18.9 & 39.4 & 57.2 & --- & --- & --- & 150\%\\
        & CosPlace & 2.5 & 10.5 & \textbf{25.0} & \textbf{4.7} & \textbf{11.2} & 22.1 & 4.3 & 9.3 & 16.1 & 2.0 & 6.0 & 11.0 & --- & --- & --- & 100\%\\
        & EigenPlaces & 2.3 & 11.0 & 25.2 & 6.7 & \textbf{15.4} & \textbf{27.1} & \textbf{7.1} & 15.2 & 22.4 & 3.5 & 9.9 & 17.8 & --- & --- & --- & 150\%\\
        & \textcolor{red}{MixVPR} & 1.1 & 8.0 & \textbf{26.1} & 4.1 & 12.5 & 25.2 & \textbf{7.2} & \textbf{14.5} & 24.2 & 5.5 & 10.6 & 17.4 & --- & --- & --- & 150\%\\
        & NetVLAD & 1.0 & 7.3 & \textbf{21.9} & \textbf{1.4} & \textbf{8.2} & 18.7 & 0.9 & 4.0 & 9.5 & 0.6 & 2.3 & 5.2 & --- & --- & --- & 100\%\\
        & SALAD & 2.1 & 14.0 & \textbf{36.4} & 4.8 & 16.0 & 33.8 & \textbf{6.9} & \textbf{17.3} & 30.0 & 5.8 & 14.1 & 24.1 & --- & --- & --- & 150\%\\
        & \textcolor{blue}{SelaVPR} & 2.3 & 16.4 & 43.5 & 9.1 & 30.6 & \textbf{55.7} & \textbf{11.9} & \textbf{35.0} & 53.6 & 8.7 & 26.3 & 44.5 & --- & --- & --- & 150\%\\
        \hline
        \hline
        \multirow{7}{*}{VPAir (ortho)} & \textcolor{blue}{AnyLoc} & 13.3 & 42.8 & 72.0 & 42.9 & 69.3 & 83.2 & 59.5 & 76.3 & \textbf{86.3} & \textbf{59.7} & \textbf{76.9} & 84.5 & --- & --- & --- & 200\%\\
        & CosPlace & 8.8 & 29.1 & 50.8 & 24.8 & 49.4 & 66.7 & \textbf{28.8} & \textbf{54.7} & \textbf{70.0} & 24.3 & 48.2 & 63.9 & --- & --- & --- & 150\%\\
        & \textcolor{red}{EigenPlaces} & 7.5 & 28.5 & 56.5 & 34.0 & 60.4 & 75.5 & \textbf{48.4} & \textbf{72.5} & \textbf{82.0} & 46.9 & 69.1 & 79.7 & --- & --- & --- & 150\%\\
        & \textcolor{green}{MixVPR} & 5.7 & 22.8 & 50.8 & 43.5 & 69.2 & 83.1 & \textbf{67.2} & \textbf{85.1} & \textbf{91.9} & 66.0 & 84.1 & 90.5 & --- & --- & --- & 150\%\\
        & NetVLAD & 2.1 & 13.1 & 39.1 &\textbf{ 9.4} & \textbf{26.3} & \textbf{47.1} & 9.2 & 22.6 & 40.7 & 6.8 & 17.1 & 30.0 & --- & --- & --- & 100\%\\
        & SALAD & 10.4 & 34.2 & 61.6 & 32.7 & 56.3 & 76.0 & \textbf{47.2} & \textbf{70.0} & \textbf{82.1} & 47.0 & 67.4 & 78.9 & --- & --- & --- & 150\%\\
        & SelaVPR & 5.1 & 21.4 & 45.6 & 18.5 & 43.9 & 66.5 & \textbf{29.5} & \textbf{53.3} & \textbf{69.0} & 23.9 & 46.6 & 63.2 & --- & --- & --- & 150\%\\
        \hline
        \hline
        \multirow{7}{*}{ALTO} & \textcolor{blue}{AnyLoc} & 3.7 & 19.1 & 40.0 & 10.7 & 30.0 & 44.3 & 19.3 & 34.1 & \textbf{48.5} & \textbf{20.0} & \textbf{35.4} & 46.5 & 18.9 & 32.6 & 43.7 & 200\%\\
        & CosPlace & 5.0 & 12.6 & \textbf{28.7} & \textbf{6.1} & \textbf{15.4} & 26.5 & 5.0 & 13.3 & 26.1 & 6.1 & 12.8 & 21.7 & 4.1 & 11.5 & 20.2 & 100\%\\
        & EigenPlaces & 4.6 & 12.0 & 31.1 & \textbf{7.6} & \textbf{21.3} & \textbf{37.6} & 7.4 & 18.9 & 35.0 & 6.5 & 16.1 & 27.6 & 4.8 & 15.7 & 27.0 & 100\%\\
        & \textcolor{green}{MixVPR} & 2.6 & 14.6 & 27.4 & 8.0 & 22.8 & 38.7 & 17.4 & 35.4 & 50.2 & \textbf{20.7} & \textbf{38.9} & \textbf{53.5} & 17.0 & 34.6 & 50.4 & 200\%\\
        & NetVLAD & 2.6 & 7.2 & 14.3 & \textbf{4.8} & \textbf{8.5} & \textbf{17.0} & 0.9 & 5.9 & 16.1 & 0.9 & 5.2 & 10.4 & 0.9 & 3.3 & 6.7 & 100\%\\
        & \textcolor{red}{SALAD} & 6.1 & 18.7 & 37.8 & 12.2 & 24.1 & 40.7 & 16.7 & 29.8 & 45.0 & \textbf{18.3} & \textbf{31.7} & \textbf{46.5} & 15.9 & 28.9 & 42.8 & 200\%\\
        & SelaVPR & 4.1 & 17.2 & 32.6 & 9.6 & 23.5 & 34.6 & 10.2 & 27.0 & 37.8 & \textbf{15.0} & \textbf{28.3} & \textbf{42.8} & 9.1 & 22.0 & 33.9 & 200\%\\
        \hline
        \hline
        \multirow{7}{*}{AMtown01} & \textcolor{blue}{AnyLoc} & 15.7 & 52.7 & 83.2 & \textbf{25.3} & \textbf{63.7} & \textbf{89.3} & 18.3 & 52.7 & 77.4 & 10.7 & 37.4 & 65.6 & --- & --- & --- & 100\%\\
        & CosPlace & \textbf{7.2} & \textbf{20.3} & \textbf{46.9} & 2.1 & 12.4 & 23.9 & 1.3 & 9.7 & 25.7 & 0.6 & 5.1 & 13.8 & --- & --- & --- & 50\%\\
        & EigenPlaces & \textbf{8.4} & \textbf{22.4} & \textbf{62.4} & 4.7 & 15.2 & 37.7 & 2.9 & 15.9 & 34.5 & 6.3 & 13.0 & 18.5 & --- & --- & --- & 50\%\\
        & MixVPR & 1.1 & 9.7 & \textbf{38.3} & \textbf{8.4} & \textbf{18.4} & 34.6 & 6.8 & 15.2 & 29.0 & 2.9 & 11.1 & 22.3 & --- & --- & --- & 100\%\\
        & NetVLAD & \textbf{5.6} & \textbf{31.1} & \textbf{64.9} & 4.5 & 18.5 & 39.0 & 0.8 & 6.8 & 23.5 & 0.1 & 2.7 & 13.8 & --- & --- & --- & 50\%\\
        & \textcolor{red}{SALAD} & 5.9 & 42.6 & \textbf{88.2} & 17.6 & \textbf{45.0} & 76.6 & \textbf{18.4} & 44.5 & 67.9 & 11.6 & 26.0 & 46.4 & --- & --- & --- & 100\%\\
        & \textcolor{green}{SelaVPR} & 4.7 & 40.6 & \textbf{91.8} & \textbf{26.7} & \textbf{62.9} & 87.3 & 16.9 & 56.8 & 81.0 & 10.9 & 32.0 & 56.5 & --- & --- & --- & 100\%\\
    \end{tabular}
    }
    \caption{Comparison of different zoom levels, {\bf{VPR Recall}} value in \%. The most favorable results in each row are emphasized in bold. The superior method for each dataset is identified in green, followed by the second-best method in blue, and the third-best method in red.}
    \label{tab:zoom}
\end{table*}

The findings suggest that the optimal zoom level is contingent not only on the characteristics of the test dataset, but also on the specific localization method employed; so this is a hyperparameter that \underline{needs to be tuned in each scenario}. Nevertheless, among the top-performing methods (AnyLoc~\cite{keetha2023anyloc}, MixVPR~\cite{ali2023mixvpr}, SALAD~\cite{izquierdo2023optimal} and SelaVPR~\cite{lu2024towards}), there is typically a convergence on the optimal zoom level within the same dataset. Subsequent experiments were conducted exclusively at the optimal zoom level identified for each algorithm.

\subsubsection{Overlap level}
All the VPR methods under consideration were utilized to investigate the impact of the overlap level between neighboring frames. Coverage levels of 0\%, 25\%, and 50\% were examined. The {\bf{VPR Recall}} metric was used. The results of comparison can be seen in Table~\ref{tab:overlap}.

\begin{table*}
    \centering
    \begin{tabular}{l|l|c|c|c|c|c|c|c|c|c}
        \multirow{2}{*}{Dataset} & Localization & \multicolumn{3}{c|}{Overlap 0\%} & \multicolumn{3}{c|}{Overlap 25\%} & \multicolumn{3}{c}{Overlap 50\%} \\\cline{3-11}
        & method & R@1 & R@10 & R@50 & R@1 & R@10 & R@50 & R@1 & R@10 & R@50 \\
        \hline \hline
        \multirow{7}{*}{VPAir (plane)} & \textcolor{green}{AnyLoc} & 12.9 & 30.3 & 48.9 & 21.7 & 43.6 & 61.2 & \textbf{26.3} & \textbf{49.2} & \textbf{66.0}\\
        & CosPlace & 2.6 & 9.3 & 19.3 & 4.7 & 11.2 & \textbf{22.1} & \textbf{4.9} & \textbf{11.6} & 21.8\\
        & EigenPlaces & 2.6 & 8.6 & 15.9 & 7.1 & 15.2 & 22.4 & \textbf{7.2} & \textbf{15.5} & \textbf{23.8}\\
        & MixVPR & 3.6 & 9.6 & 18.9 & \textbf{7.2} & 14.5 & \textbf{24.2} & 6.9 & \textbf{14.8} & \textbf{24.2}\\
        & NetVLAD & 1.4 & 6.9 & 17.3 & 1.4 & \textbf{8.2} & \textbf{18.7} & \textbf{1.7} & 6.6 & 15.9\\
        & \textcolor{red}{SALAD} & 4.7 & 12.8 & 24.9 & 6.9 & 17.3 & 30.0 & \textbf{9.3} & \textbf{21.7} & \textbf{33.3}\\
        & \textcolor{blue}{SelaVPR} & 7.8 & 24.7 & 44.0 & 11.9 & 35.0 & 53.6 & \textbf{13.6} & \textbf{37.3} & \textbf{57.8}\\
        \hline \hline
        \multirow{7}{*}{VPAir (ortho)} & \textcolor{blue}{AnyLoc} & 44.1 & 64.6 & 76.0 & 59.7 & 76.9 & 84.5 & \textbf{69.1} & \textbf{81.1} & \textbf{86.8}\\
        & CosPlace & 21.8 & 41.8 & 57.0 & 28.8 & 54.7 & 70.0 & \textbf{39.0} & \textbf{60.3} & \textbf{73.4}\\
        & EigenPlaces & 34.3 & 53.8 & 66.7 & 48.4 & 72.5 & 82.0 & \textbf{60.5} & \textbf{77.5} & \textbf{84.8}\\
        & \textcolor{green}{MixVPR} & 45.5 & 65.7 & 77.8 & 67.2 & 85.1 & 91.9 & \textbf{78.9} & \textbf{90.7} & \textbf{94.5}\\
        & NetVLAD & 7.9 & 22.1 & 40.0 & 9.4 & 26.3 & \textbf{47.1} & \textbf{11.6} & \textbf{28.9} & 45.8\\
        & \textcolor{red}{SALAD} & 31.1 & 50.8 & 64.9 & 47.2 & 70.0 & 82.1 & \textbf{65.7} & \textbf{80.7} & \textbf{88.5}\\
        & SelaVPR & 20.4 & 40.2 & 56.9 & 29.5 & 53.3 & 69.0 & \textbf{37.8} & \textbf{60.3} & \textbf{74.4}\\
        \hline \hline
        \multirow{7}{*}{ALTO} & \textcolor{blue}{AnyLoc} & 14.6 & 28.7 & 39.6 & 20.0 & 35.4 & 46.5 & \textbf{25.7} & \textbf{39.1} & \textbf{48.0}\\
        & CosPlace & 4.1 & 10.9 & 22.0 & 6.1 & 15.4 & 26.5 & \textbf{7.2} & \textbf{16.3} & \textbf{28.0}\\
        & EigenPlaces & 7.0 & 17.4 & 32.2 & 7.6 & 21.3 & \textbf{37.6} & \textbf{11.1} & \textbf{23.7} & 36.1\\
        & \textcolor{green}{MixVPR} & 13.5 & 24.8 & 40.7 & 20.7 & 38.9 & 53.5 & \textbf{32.2} & \textbf{49.3} & \textbf{61.1}\\
        & NetVLAD & 2.6 & 9.6 & 17.0 & \textbf{4.8} & 8.5 & 17.0 & 4.6 & \textbf{11.7} & \textbf{20.2}\\
        & \textcolor{red}{SALAD} & 9.8 & 20.9 & 35.4 & 18.3 & 31.7 & 46.5 & \textbf{23.9} & \textbf{40.9} & \textbf{56.7}\\
        & SelaVPR & 8.7 & 22.4 & 33.0 & 15.0 & 28.3 & 42.8 & \textbf{16.3} & \textbf{31.5} & \textbf{42.8}\\
        \hline \hline
        \multirow{7}{*}{AMtown01} & \textcolor{green}{AnyLoc} & 18.3 & 55.8 & 82.6 & 25.3 & 63.7 & \textbf{89.3} & \textbf{31.6} & \textbf{67.6} & 87.1\\
        & CosPlace & 2.5 & 15.3 & 44.8 & 7.2 & \textbf{20.3} & \textbf{46.9} & \textbf{8.7} & 19.0 & 28.5\\
        & EigenPlaces & 2.7 & 18.3 & \textbf{64.3} & \textbf{8.4} & \textbf{22.4} & 62.4 & 6.8 & 20.3 & 40.6\\
        & MixVPR & 6.3 & 12.6 & 28.4 & \textbf{8.4} & \textbf{18.4} & \textbf{34.6} & \textbf{8.4} & 17.1 & 29.6\\
        & NetVLAD & 4.0 & 21.8 & 60.5 & 5.6 & 31.1 & \textbf{64.9} & \textbf{6.5} & \textbf{32.5} & 60.4\\
        & \textcolor{red}{SALAD} & 11.6 & 34.0 & 61.8 & 17.6 & \textbf{45.0} & \textbf{76.6} & \textbf{19.2} & \textbf{45.0} & 72.1\\
        & \textcolor{blue}{SelaVPR} & 17.2 & 54.7 & 85.5 & \textbf{26.7} & 62.9 & 87.3 & 25.6 & \textbf{69.4} & \textbf{91.0}\\
    \end{tabular}
    \caption{Comparison of different overlap levels, {\bf{VPR Recall}} value in \%. The most favorable results in each row are emphasized in bold. The superior method for each dataset is identified in green, followed by the second-best method in blue, and the third-best method in red.}
    \label{tab:overlap}
\end{table*}

The findings indicate that \underline{increasing the level of overlap} between frames \underline{enhances the efficacy} of global localization systems, with the principal constraint in such scenarios revolving around the temporal and memory resources demanded for database construction. Upon comparative analysis, AnyLoc~\cite{keetha2023anyloc}, MixVPR~\cite{ali2023mixvpr}, SALAD~\cite{izquierdo2023optimal} and SelaVPR~\cite{lu2024towards} methods perform the best, albeit subject to variability contingent upon dataset. In further experiments, \underline{we used an overlap level equal to 50\%}.

\subsubsection{Re-ranking}
We conducted a comparative analysis of re-ranking techniques using the top-100 predictions produced by diverse global localization methods. For global localization, we employed AnyLoc~\cite{keetha2023anyloc}, MixVPR~\cite{ali2023mixvpr}, SALAD~\cite{izquierdo2023optimal} and SelaVPR~\cite{lu2024towards}, which demonstrated superior performance compared to other methods. For re-ranking, we used all the methods considered: SuperGlue~\cite{sarlin2020superglue}, LightGlue~\cite{lindenberger2023lightglue} and SelaVPR~\cite{lu2024towards}. Notably, SelaVPR, which integrates both global localization and re-ranking steps, was exclusively evaluated in conjunction with itself. The {\bf{VPR Recall}} metric was used. The outcomes of this comparison are detailed in Table~\ref{tab:reranking}.

\begin{table*}
    \centering
    \begin{tabular}{l|l|c|c|c|c|c|c|c|c|c|c|c|c}
        \multirow{2}{*}{Dataset} & Localization & \multicolumn{3}{c|}{W/o re-ranking} & \multicolumn{3}{c|}{SuperGlue} & \multicolumn{3}{c|}{LightGlue} & \multicolumn{3}{c}{SelaVPR} \\\cline{3-14}
        & method & R@1 & R@5 & R@10 & R@1 & R@5 & R@10 & R@1 & R@5 & R@10 & R@1 & R@5 & R@10 \\
        \hline \hline
        \multirow{4}{*}{VPAir (plane)} & AnyLoc & 26.3 & 41.9 & \textcolor{green}{49.2} & 23.7 & 32.6 & 37.7 & 17.1 & 23.5 & 28.4 & --- & --- & --- \\
        & MixVPR & 6.9 & 11.4 & 14.8 & 10.8 & 14.7 & 17.0 & 8.8 & 11.8 & 14.2 & --- & --- & --- \\
        & SALAD & 9.3 & 17.3 & 21.7 & 17.9 & 22.2 & 24.6 & 13.5 & 16.5 & 19.3 & --- & --- & --- \\
        & \textcolor{green}{SelaVPR} & 13.6 & 29.7 & 37.3 & --- & --- & --- & --- & --- & --- & \textcolor{green}{26.9} & \textcolor{green}{42.8} & \textcolor{green}{49.2} \\
        \hline \hline
        \multirow{4}{*}{VPAir (ortho)} & AnyLoc & 69.1 & 78.2 & 81.1 & 72.7 & 80.9 & 82.8 & 71.6 & 79.7 & 81.3 & --- & --- & --- \\
        & \textcolor{green}{MixVPR} & 78.9 & \textcolor{green}{87.8} & \textcolor{green}{90.7} & 78.5 & 84.0 & 86.0 & \textcolor{green}{79.9} & 85.4 & 88.3 & --- & --- & --- \\
        & SALAD & 65.7 & 77.4 & 80.7  & 75.1 & 80.5 & 82.0 & 75.5 & 79.5 & 81.9 & --- & --- & --- \\
        & SelaVPR & 37.8 & 54.5 & 60.3 & --- & --- & --- & --- & --- & --- & 68.3 & 74.1 & 76.0 \\
        \hline \hline
        \multirow{4}{*}{ALTO} & AnyLoc & 25.7 & 34.3 & 39.1 & 35.2 & 40.4 & 41.7 & 35.7 & 39.8 & 42.6 & --- & --- & --- \\
        & \textcolor{green}{MixVPR} & 32.2 & 42.6 & 49.3 & 38.9 & 44.6 & 47.8 & \textcolor{green}{41.7} & \textcolor{green}{47.2} & \textcolor{green}{50.2} & --- & --- & --- \\
        & SALAD & 23.9 & 35.9 & 40.9 & 41.1 & 46.1 & 48.3 & 40.4 & 45.2 & 48.0 & --- & --- & --- \\
        & SelaVPR & 16.3 & 26.5 & 31.5 & --- & --- & --- & --- & --- & --- & 33.3 & 42.6 & 44.8 \\
        \hline \hline
        \multirow{4}{*}{AMtown01} & AnyLoc & 31.6 & 56.5 & 67.6 & 38.4 & 43.7 & 49.7 & 36.8 & 40.2 & 43.9 & --- & --- & --- \\
        & MixVPR & 8.4 & 13.4 & 17.1 & 23.9 & 25.4 & 26.8 & 23.7 & 25.3 & 26.6 & --- & --- & --- \\
        & SALAD & 19.2 & 35.8 & 45.0 & 37.1 & 41.1 & 45.0 & 36.6 & 39.0 & 41.5 & --- & --- & --- \\
        & \textcolor{green}{SelaVPR} & 25.6 & 57.8 & 69.4 & --- & --- & --- & --- & --- & --- & \textcolor{green}{45.5} & \textcolor{green}{69.6} & \textcolor{green}{78.7} \\
    \end{tabular}
    \caption{Comparison of different re-ranking methods, {\bf{VPR Recall}} value in \%. Top-100 candidates are used everywhere. The best configurations for R@1, R@5 and R@10 are highlighted in green.}
    \label{tab:reranking}
\end{table*}

The results underscore the variability in the optimal configuration of VPR, encompassing global localization and re-ranking, across distinct datasets. Notably, for the VPAir plane~\cite{schleiss2022vpair} and AMtown01~\cite{li2024mars} sequences, the combined SelaVPR~\cite{lu2024towards} method demonstrates superior performance, contrasting sharply with its relatively poorer performance on the other two sequences. Conversely, for the VPAir ortho~\cite{schleiss2022vpair} sequence, the MixVPR~\cite{ali2023mixvpr} without re-ranking step yields the most favorable outcomes, while on the ALTO~\cite{cisneros2022alto} dataset this method shows itself better in conjunction with LightGlue~\cite{lindenberger2023lightglue}.

\subsubsection{Local alignment}
We employed the AnyLoc~\cite{keetha2023anyloc}, MixVPR~\cite{ali2023mixvpr}, SALAD~\cite{izquierdo2023optimal} and SelaVPR~\cite{lu2024towards} methods, along with all re-ranking methods, to estimate the entire visual geolocalization pipeline, including the local alignment step. SelaVPR was only used in conjunction with itself. The {\bf{Georeference Recall}} metric was utilized, with thresholds set at 10, 50, and 100 meters. The outcomes of this evaluation are summarized in Table~\ref{tab:reference}.

\begin{table*}
    \centering
    \begin{tabular}{l|l|c|c|c|c|c|c|c|c|c}
        \multirow{2}{*}{Dataset} & Localization & \multicolumn{3}{c|}{SuperGlue} & \multicolumn{3}{c|}{LightGlue} & \multicolumn{3}{c}{SelaVPR} \\\cline{3-11}
        & method & 10m & 50m & 100m & 10m & 50m & 100m & 10m & 50m & 100m \\
        \hline \hline
        \multirow{4}{*}{VPAir (plane)} & AnyLoc & \textcolor{green}{4.6} & \textcolor{green}{15.4} & \textcolor{green}{18.1} & 3.7 & 11.6 & 13.9 & --- & --- & --- \\
        & MixVPR & 2.6 & 7.4 & 8.5 & 2.4 & 6.5 & 7.6 & --- & --- & --- \\
        & SALAD & 4.0 & 12.9 & 14.7 & 3.2 & 10.3 & 11.3 & --- & --- & --- \\
        & SelaVPR & --- & --- & --- & --- & --- & --- & 1.1 & 9.2 & 15.7 \\
        \hline \hline
        \multirow{4}{*}{VPAir (ortho)} & AnyLoc & 52.1 & 71.8 & 74.7 & 49.9 & 68.7 & 72.2 & --- & --- & --- \\
        & MixVPR & \textcolor{green}{58.3} & \textcolor{green}{74.8} & \textcolor{green}{77.2} & 57.4 & 73.9 & 76.8 & --- & --- & --- \\
        & SALAD & 56.9 & 71.8 & 74.0 & 55.4 & 70.8 & 73.4 & --- & --- & --- \\
        & SelaVPR & --- & --- & --- & --- & --- & --- & 16.0 & 57.5 & 65.0 \\
        \hline \hline
        \multirow{4}{*}{ALTO} & AnyLoc & 5.2 & 28.7 & 33.0 & 3.9 & 28.0 & 31.3 & --- & --- & --- \\
        & MixVPR & 5.9 & 33.3 & 38.0 & 5.0 & \textcolor{green}{34.8} & \textcolor{green}{38.5} & --- & --- & --- \\
        & SALAD & \textcolor{green}{6.5} & 34.1 & 37.6 & 4.1 & 33.0 & 36.1 & --- & --- & --- \\
        & SelaVPR & --- & --- & --- & --- & --- & --- & 1.7 & 20.9 & 29.6 \\
        \hline \hline
        \multirow{4}{*}{AMtown01} & AnyLoc & 5.3 & \textcolor{green}{37.7} & 38.8 & 6.0 & 36.5 & 37.1 & --- & --- & --- \\
        & MixVPR & 4.8 & 25.2 & 25.6 & 5.3 & 23.5 & 24.0 & --- & --- & --- \\
        & SALAD & 5.8 & 36.9 & 37.8 & 6.1 & 36.3 & 36.6 & --- & --- & --- \\
        & SelaVPR & --- & --- & --- & --- & --- & --- & \textcolor{green}{7.4} & 35.4 & \textcolor{green}{46.0} \\
    \end{tabular}
    \caption{Comparison of VPR methods with local alignment step, {\bf{Georeference Recall}} value in \%. The best configurations for each threshold are highlighted in green.}
    \label{tab:reference}
\end{table*}

The results indicate that on the VPAir plane~\cite{schleiss2022vpair} dataset, the AnyLoc~\cite{keetha2023anyloc} method combined with SuperGlue~\cite{sarlin2020superglue} performs optimally. Similarly, on the VPAir ortho~\cite{schleiss2022vpair} dataset, the MixVPR~\cite{ali2023mixvpr} method, also combined with SuperGlue, yields the best performance. On the ALTO~\cite{cisneros2022alto} sequence, the highest performance varies with the threshold value, with SALAD~\cite{izquierdo2023optimal} and MixVPR demonstrating superior results when paired with SuperGlue and LightGlue, respectively. For the AMtown01 dataset, the combined SelaVPR~\cite{lu2024towards} approach predominantly achieves the best results.

\subsection{Performance evaluation}

\subsubsection{Time measurements}
For the time measurements, the same test bench utilized for the \emph{Local alignment} step estimation was employed. All measurements were conducted on the following machine configuration: AMD Ryzen Threadripper 3970X, NVIDIA GeForce RTX 3090, and 128GB RAM. Time measurements were taken for each stage of the visual geolocalization process, specifically: global descriptor extraction, database search, local feature extraction, re-ranking, and local alignment. Since both LightGlue~\cite{lindenberger2023lightglue} and SuperGlue~\cite{sarlin2020superglue} use SuperPoint~\cite{detone2018superpoint} features, the results for these methods for the local feature extraction step are combined. The results can be seen in Table~\ref{tab:perf}.

\begin{table*}
    \centering
    \begin{tabular}{l|l|c|c|c|c|c|c|c|c|c|c}
         \multirow{3}{*}{Dataset} & \multirow{3}{*}{\shortstack{Localization \\ method}} & \multirow{3}{*}{\shortstack{Descriptor \\ calculation}} & \multirow{3}{*}{\shortstack{Database \\ search}} & \multicolumn{2}{c|}{Local features} & \multicolumn{3}{c|}{\multirow{2}{*}{Re-ranking}} & \multicolumn{3}{c}{Local}\\
         & ~ & ~ & ~ & \multicolumn{2}{c|}{calculation} & \multicolumn{3}{c|}{~} & \multicolumn{3}{c}{alignment}\\\cline{5-12}
         & ~ & ~ & ~ & SP & Sela & SG & LG & Sela & SG & LG & Sela \\
         \hline \hline
         \multirow{4}{*}{VPAir (plane)} & AnyLoc & 0.63 & 1.51 & \textcolor{green}{\multirow{4}{*}{0.04}} & \textcolor{green}{\multirow{4}{*}{0.04}} & 22.86 & 1.04 & --- & \textcolor{green}{0.07} & 0.11 & ---\\
         & MixVPR & 0.08 & 0.12 & ~ & ~ & 23.06 & 1.03 & --- & 0.09 & 0.12 & ---\\
         & SALAD & 0.09 & 0.26 & ~ & ~ & 23.44 & 1.08 & --- & \textcolor{green}{0.07} & 0.09 & ---\\
         & SelaVPR & \textcolor{green}{0.04} & \textcolor{green}{0.03} & ~ & ~ & --- & --- & \textcolor{green}{0.08} & --- & --- & \textcolor{green}{0.07}\\
         \hline \hline
         \multirow{4}{*}{VPAir (ortho)} & AnyLoc & 0.63 & 2.68 & \textcolor{green}{\multirow{4}{*}{0.04}} & \textcolor{green}{\multirow{4}{*}{0.04}} & 23.64 & 1.08 & --- & 0.05 & 0.07 & ---\\
         & MixVPR & 0.08 & 0.12 & ~ & ~ & 23.11 & 1.08 & --- & \textcolor{green}{0.04} & 0.07 & ---\\
         & SALAD & 0.09 & 0.26 & ~ & ~ & 23.57 & 1.14 & --- & \textcolor{green}{0.04} & 0.05 & ---\\
         & SelaVPR & \textcolor{green}{0.04} & \textcolor{green}{0.03} & ~ & ~ & --- & --- & \textcolor{green}{0.08} & --- & --- & 0.06\\
         \hline \hline
         \multirow{4}{*}{ALTO} & AnyLoc & 0.37 & 1.32 & \textcolor{green}{\multirow{4}{*}{0.04}} & \textcolor{green}{\multirow{4}{*}{0.04}} & 24.08 & 1.16 & --- & \textcolor{green}{0.06} & 0.07 & ---\\
         & MixVPR & 0.08 & 0.11 & ~ & ~ & 17.28 & 1.18 & --- & 0.07 & 0.10 & ---\\
         & SALAD & 0.05 & 0.21 & ~ & ~ & 17.87 & 1.05 & --- & 0.08 & 0.10 & ---\\
         & SelaVPR & \textcolor{green}{0.04} & \textcolor{green}{0.03} & ~ & ~ & --- & --- & \textcolor{green}{0.09} & --- & --- & 0.09\\
         \hline \hline
         \multirow{4}{*}{AMtown01} & AnyLoc & 0.84 & 0.44 & \textcolor{green}{\multirow{4}{*}{0.14}} & \multirow{4}{*}{0.17} & 18.59 & 1.07 & --- & 0.13 & 0.14 & ---\\
         & MixVPR & 0.20 & 0.04 & ~ & ~ & 15.67 & 1.05 & --- & 0.15 & 0.15 & ---\\
         & SALAD & 0.24 & 0.07 & ~ & ~ & 18.29 & 1.06 & --- & 0.13 & 0.14 & ---\\
         & SelaVPR & \textcolor{green}{0.17} & \textcolor{green}{0.01} & ~ & ~ & --- & --- & \textcolor{green}{0.09} & --- & --- & \textcolor{green}{0.02}\\
    \end{tabular}
    \caption{Performance of VPR methods, time in seconds. The best result for each step is highlighted in green.}
    \label{tab:perf}
\end{table*}

The comparison findings indicate that computing descriptors and conducting a database search are relatively swift across all methods, except for AnyLoc~\cite{keetha2023anyloc}, where these tasks significantly prolong. This discrepancy arises from its utilization of a large ViT-G DINOv2~\cite{oquab2023dinov2} model and the fact that AnyLoc generates descriptors of high dimensionality (49152). Local features computation and alignment demonstrate rapidity across all methods. Notably, there's a discernible contrast in the runtime of re-ranking methods. While SelaVPR~\cite{lu2024towards} completes in less than 0.1 seconds, LightGlue~\cite{lindenberger2023lightglue} demands around 1 second. SuperGlue~\cite{sarlin2020superglue}, on the other hand, operates at an even slower pace, necessitating 15 to 25 seconds, contingent upon the dataset.

\subsubsection{Memory consumption}
To compare the memory consumption of different methods, the corresponding descriptors for AnyLoc~\cite{keetha2023anyloc}, MixVPR~\cite{ali2023mixvpr}, SALAD~\cite{izquierdo2023optimal} and SelaVPR~\cite{lu2024towards} methods and local features for SuperGlue~\cite{sarlin2020superglue}, LightGlue~\cite{lindenberger2023lightglue} and SelaVPR were saved to a hard drive. The optimal zoom level was employed for each method, with an overlap level set to 50\%. Consequently, for the VPAir~\cite{schleiss2022vpair} dataset, statistics are provided at a zoom level of 200\% exclusively for the AnyLoc method. The ensuing outcomes are accessible for scrutiny in Table~\ref{tab:memory}.


\begin{table}
    \centering
    \resizebox{\columnwidth}{!}{
    \begin{tabular}{l|l|c|c|c|c}
         \multicolumn{2}{l|}{Dataset} & \multicolumn{2}{c|}{VPAir} & ALTO & AMtown01 \\
         \hline \hline
         \multicolumn{2}{l|}{Zoom level} & 150\% & 200\% & 200\% & 100\% \\
         \hline \hline
         \multirow{4}{*}{\shortstack{Descriptors \\ size}} & AnyLoc & 7.9 Gb & 13.9 Gb & 6.4 Gb & 2.3 Gb \\
         & MixVPR & 0.7 Gb & --- & 0.5 Gb & 0.2 Gb \\
         & SALAD & 1.4 Gb & --- & 1.1 Gb & 0.4 Gb \\
         & SelaVPR & \textcolor{green}{0.2 Gb} & --- & \textcolor{green}{0.1 Gb} & \textcolor{green}{< 0.1 Gb} \\
         \hline \hline
         \multirow{2}{*}{\shortstack{Local features \\ size}} & SP & \textcolor{green}{30.7 Gb} & 36.0 Gb & \textcolor{green}{16.2 Gb} & \textcolor{green}{7.8 Gb} \\
         & SelaVPR & 75.7 Gb & --- & 61.4 Gb & 22.1 Gb \\
    \end{tabular}}
    \caption{Memory consumption of VPR methods. The best results for each dataset are highlighted in green.}
    \label{tab:memory}
\end{table}

It's important to highlight that AnyLoc's~\cite{keetha2023anyloc} descriptors require significantly more memory compared to descriptors in other methods due to their high dimensionality. Similarly, SelaVPR's~\cite{lu2024towards} local features demand substantially more memory than the SuperPoint~\cite{detone2018superpoint} local features utilized by LightGlue~\cite{lindenberger2023lightglue} and SuperGlue~\cite{sarlin2020superglue}. However, these features also consume a considerable amount of memory, particularly noticeable with large map sizes, presenting potential challenges for devices with limited memory capacity.

\subsection{Discussion}
The culmination of the experimental investigation yields several significant conclusions. Firstly, the selection of an appropriate \textbf{zoom level} stands as a crucial factor in augmenting the efficacy of any VPR system. When aiming to localize images captured at varying altitudes, it is advisable to utilize multiple databases constructed at diverse zoom levels. Moreover, augmenting the \textbf{overlap level} predominantly enhances the quality of VPR methods.

Secondly, upon scrutinizing various \textbf{VPR techniques} alongside diverse \textbf{re-ranking strategies}, discerning an optimal configuration proves arduous. The efficacy of different methods appears contingent upon the specific characteristics of the test data. This trend persists when evaluating VPR system quality in conjunction with the \textbf{Local alignment} step, where the ranking of methods hinges on the dataset under examination.

In terms of \textbf{temporal} and \textbf{memory} metrics, a clearer pattern emerges. SelaVPR~\cite{lu2024towards} emerges as the swiftest method across most stages, although certain other methodologies demonstrate comparable speeds at select stages. Additionally, SelaVPR demands the least memory allocation for storing global descriptors among all methods. However, SelaVPR necessitates a higher memory allocation for storing local features compared to SuperPoint~\cite{detone2018superpoint}, the feature descriptor employed by SuperGlue~\cite{sarlin2020superglue} and LightGlue~\cite{lindenberger2023lightglue}. When addressing time and memory constraints in the scaling of VPR methods, the principal challenge is the substantial memory consumption associated with local features. For large-scale maps, a feasible solution is to use global localization methods without re-ranking step. Among the VPR methods we assessed, all are potentially suitable for this approach, with the exception of AnyLoc, which is excluded due to the high dimensionality of its descriptor.

It is important to acknowledge that while a few visual geolocalization techniques demonstrate frame rates of 1 FPS or higher, which may be adequate for various practical applications, our experiments were conducted using a high-performance computing system. Consequently, evaluating the performance of these methods on microcomputers integrated into airborne devices represents a logical extension of our research.